\documentclass[10pt,twocolumn,letterpaper]{article}

\usepackage{multirow}
\usepackage{cvpr}
\usepackage{times}
\usepackage{epsfig}
\usepackage{graphicx}
\usepackage{amsmath}
\usepackage{amssymb}
\usepackage[export]{adjustbox}
\usepackage[numbers,sort]{natbib} 

\usepackage{multido}
\usepackage{subcaption}
\usepackage{comment}

\newcommand\clipexamples[1]{%
  \multido{\I=1+1}{6}{%
    \includegraphics[width=0.12\textwidth]{clip_examples/#1_\I.jpg}
    }}

\usepackage[pagebackref=true,breaklinks=true,letterpaper=true,colorlinks,bookmarks=false]{hyperref}

\cvprfinalcopy

\ifcvprfinal\fi
\begin{document}

\title{The Kinetics Human Action Video Dataset}

\author{Will Kay\\
{\tt\small wkay@google.com} \\
\and
Jo{\~a}o Carreira \\
{\tt\small joaoluis@google.com} \\
\and
Karen Simonyan \\
{\tt\small simonyan@google.com} \\
\and
Brian Zhang \\
{\tt\small brianzhang@google.com} \\
\and
Chloe Hillier \\
{\tt\small chillier@google.com} \\
\and
Sudheendra Vijayanarasimhan \\
{\tt\small svnaras@google.com} \\
\and
Fabio Viola \\
{\tt\small fviola@google.com} \\
\and
Tim Green \\
{\tt\small tfgg@google.com} \\
\and
Trevor Back \\
{\tt\small back@google.com} \\
\and
Paul Natsev \\
{\tt\small natsev@google.com} \\
\and
Mustafa Suleyman \\
{\tt\small mustafasul@google.com} \\
\and
Andrew Zisserman \\
{\tt\small zisserman@google.com} \\
}

\maketitle

\begin{abstract}
We describe the DeepMind Kinetics human action video dataset.
The dataset contains 400 human action classes, with at least 400 video clips for each action. Each clip lasts around 10s and is taken from a different YouTube video. The actions are human focussed and cover a broad range of classes including human-object interactions such as playing instruments, as well as human-human interactions such as shaking hands.
We describe the statistics of the dataset, how it was collected, and give some baseline performance figures for neural network architectures trained and tested for human action classification on this dataset. We also carry out a preliminary analysis of whether imbalance in the dataset leads to bias in the classifiers.
\end{abstract}

\section{Introduction}

In this paper we introduce a new, large, video dataset for human
action classification. We developed this dataset principally because
there is a lack of such datasets for human action classification, and
we believe that having one will facilitate research in this area --
both because the dataset is large enough to train deep networks from
scratch, and also because the dataset is challenging enough to act as
a performance benchmark where the advantages of different
architectures can be teased apart.

Our aim is to provide a large scale high quality dataset, covering
a diverse range of human actions, that can be 
used for human action {\em classification}, rather than 
temporal localization. Since the use case is classification, only short clips of around
10s containing the action are included, and there are no untrimmed
videos. However, the clips also contain sound so the dataset can
potentially be used for many purposes, including multi-modal analysis.
Our inspiration in providing a dataset for classification is
ImageNet~\cite{Russakovsky15}, where the significant benefits of first training
deep networks on this dataset for classification, and then using the
trained network for other purposes (detection, image segmentation,
non-visual modalities (e.g.\ sound, depth), etc) are well known.

The Kinetics dataset can be seen as the successor to the two human
action video datasets that have emerged as the standard benchmarks for
this area: HMDB-51~\cite{Kuehne11} and
UCF-101~\cite{soomro2012ucf101}. These datasets have served the
community very well, but their usefulness is now expiring. This is
because they are simply not large enough or have sufficient variation
to train and test the current generation of human action
classification models based on deep learning. Coincidentally, one of the
motivations for introducing the HMDB dataset was that the then current
generation of action datasets was too small. The increase then was from 10 to 51
classes, and we in turn increase this to 400 classes.

Table~\ref{tab:dataset_stats} compares the size of Kinetics to a
number of recent human action datasets.  In terms of variation,
although the UCF-101 dataset contains 101 actions with 100+ clips for
each action, all the clips are taken from only 2.5k distinct videos.
For example there are 7 clips from one video of the same person
brushing their hair. This means that there is far less variation than
if the action in each clip was performed by a different person (and
different viewpoint, lighting, etc). This problem is avoided in
Kinetics as each clip is taken from a different video.

\begin{table*}[t]
\centering
\begin{tabular}{| c | r | r | r | r | r |}
  \hline
  \textbf{Dataset} & \textbf{Year} & \textbf{Actions} & \textbf{Clips} 
& \textbf{Total} & \textbf{Videos}  \\ \hline 
HMDB-51~\cite{Kuehne11} & 2011  & 51  & min 102 & 6,766 & 3,312 \\ 
UCF-101~\cite{soomro2012ucf101} & 2012  & 101  & min 101 & 13,320 & 2,500 \\ 
ActivityNet-200~\cite{caba2015activitynet} & 2015  & 200  & avg 141 & 28,108 & 19,994 \\ 
Kinetics & 2017  &  400 & min 400 & 306,245 & 306,245 \\ \hline
\end{tabular} 
\vspace{5pt}
\caption{Statistics for recent human action recognition datasets.
`Actions', specifies the number of action classes; `Clips', the number of clips per class; `Total', is the total number of clips; and `Videos', the total number of videos from which these clips are
extracted.}
\label{tab:dataset_stats}
\end{table*}

The clips are sourced from YouTube videos. Consequently, for the most part, they are not 
professionally videoed and edited material (as in TV and film videos). There can be considerable 
camera motion/shake, illumination variations, shadows, background clutter, etc. More importantly, there
are a great variety of performers (since each clip is from a different video) with differences in
{\em how} the action is performed (e.g.\ its speed), clothing, body pose and shape, age, and camera framing and viewpoint.


Our hope is that the dataset will enable a new generation of
neural network architectures to be developed for video.  For
example, architectures including multiple streams of information
(RGB/appearance, optical flow, human pose, object category recognition),
architectures using attention, etc.  That will enable the virtues (or
otherwise) of the new architectures to be demonstrated. Issues such as the tension between static and motion prediction, and 
the open question of the best method of temporal aggregation in video
(recurrent vs convolutional) may finally be resolved.

The rest of the paper is organized as: Section~\ref{dataset} gives an overview of the new
dataset; Section~\ref{collection} describes how it was collected and discusses possible imbalances in the data and their consequences for classifier bias. 
Section~\ref{baselines} gives the performance of a number of ConvNet
architectures that are trained and tested on the dataset.
Our companion paper~\cite{Carreira17} explores the benefit of pre-training an action classification network on Kinetics, and then using the
features from the network for action classification on other (smaller) datasets.

The URLs of the YouTube videos and temporal intervals of the dataset can be obtained from \url{http://deepmind.com/kinetics}.

\section{An Overview of the Kinetics Dataset \label{dataset}}

\paragraph{Content:} The dataset is focused on human actions (rather than activities or
events). The list of action classes covers: {\em Person Actions
(singular)}, e.g.\ drawing, drinking, laughing, pumping fist; {\em
Person-Person Actions}, e.g.\ hugging, kissing, shaking hands; and,
{\em Person-Object Actions}, e.g.\ opening present, mowing lawn, washing dishes.  Some actions are fine grained and require temporal reasoning
to distinguish, for example different types of swimming. Other actions
require more emphasis on the object to distinguish, for example
playing different types of wind instruments.  

There is not a deep hierarchy,
but instead there are several (non-exclusive) parent-child groupings, e.g.\
Music (playing drums, trombone, violin, \ldots);
Personal Hygiene (brushing teeth, cutting nails, washing hands, \ldots);
Dancing (ballet, macarena, tap, \ldots);
Cooking (cutting, frying, peeling, \ldots).
The full list of classes is given in the appendix, together with parent-child groupings.
Figure~\ref{fig:samples} shows clips from a sample of classes.

\paragraph{Statistics:} The dataset has 400 human action
classes, with 400--1150 clips for each action, each from a unique
video. Each clip lasts around 10s. The current version has 306,245 videos, and is divided into three
splits, one for training having 250--1000 videos per class, one for
validation with 50 videos per class and one for testing with 100 videos
per class. The statistics are given in table~\ref{tab:kinetics_stats}.
The clips are from YouTube videos and 
have a variable resolution and frame rate.

\begin{table}[ht]
\centering
\begin{tabular}{| c | c | c | }
  \hline
  \textbf{Train} & \textbf{Validation} & \textbf{Test}  \\ \hline 
250--1000 & 50  & 100  \\ \hline
\end{tabular} 
\vspace{5pt}
\caption{Kinetics Dataset Statistics. The number of clips for each class in the train/val/test partitions.}
\label{tab:kinetics_stats}
\end{table}

\paragraph{Non-exhaustive annotation.} Each class contains clips illustrating that action. However, a particular clip can
contain several actions. Interesting examples in the dataset include: ``texting" while ``driving a car"; 
``Hula hooping" while ``playing ukulele";
``brushing teeth" while ``dancing'' (of some type). In each case both of the actions are Kinetics classes, and the clip will probably only appear under only one of these classes not both, i.e.\ clips do not have complete (exhaustive) annotation.
For this reason when evaluating classification performance, a top-5 measure is more suitable than top-1. This is
similar to the situation in ImageNet~\cite{Russakovsky15}, where one of the reasons for using a top-5 measure is that images
are only labelled for a single class, although it may contain  multiple classes.

\begin{figure*}[ht]
\centering

\begin{subfigure}[t]{0.5\textwidth}
\centering
\clipexamples{headbanging/3D2Eqc7Bqgg_12_22}\\
\clipexamples{headbanging/Ob8CRCWPcdQ_2_12}\\
\caption{headbanging}
\end{subfigure}%
~
\begin{subfigure}[t]{0.5\textwidth}
\centering
\clipexamples{stretching_leg/BNdWsA_Sxns_483_493}\\
\clipexamples{stretching_leg/ezWLnTzJCWI_138_148}\\
\caption{stretching leg}
\end{subfigure}

\begin{subfigure}[t]{0.5\textwidth}
\centering
\clipexamples{shakinghands/D1_dQGxsAlI_1_11}\\
\clipexamples{shakinghands/jQsvNVnpVlg_4_14}\\
\caption{shaking hands}
\end{subfigure}%
~
\begin{subfigure}[t]{0.5\textwidth}
\centering
\clipexamples{tickling/bJaJrsoeP8w_775_785}\\
\clipexamples{tickling/bT14QoGuPcw_58_68}\\
\caption{tickling}
\end{subfigure}

\begin{subfigure}[t]{0.5\textwidth}
\centering
\clipexamples{robot_dancing/Cx6XQIwOqyM_2_12}\\
\clipexamples{robot_dancing/XUphJ8I0hC0_84_94}\\
\caption{robot dancing}
\end{subfigure}%
~
\begin{subfigure}[t]{0.5\textwidth}
\centering
\clipexamples{salsa_dancing/5Zt0gmjePzg_109_119}\\
\clipexamples{salsa_dancing/VvbjmB4ss34_94_104}\\
\caption{salsa dancing}
\end{subfigure}

\begin{subfigure}[t]{0.5\textwidth}
\centering
\clipexamples{riding_a_bike/2apQFICLqLk_1_11}\\
\clipexamples{riding_a_bike/JJLqJhgJNNg_1_11}\\
\caption{riding a bike}
\end{subfigure}%
~
\begin{subfigure}[t]{0.5\textwidth}
\centering
\clipexamples{riding_unicycle/C3j4HFUUiQE_48_58}\\
\clipexamples{riding_unicycle/DCMjjC63AD4_31_41}\\
\caption{riding unicycle}
\end{subfigure}

\begin{subfigure}[t]{0.5\textwidth}
\centering
\clipexamples{playing_violin/CD6YX1whJG0_96_106}\\
\clipexamples{playing_violin/rzElVXuj-sQ_98_108}\\
\caption{playing violin}
\end{subfigure}%
~
\begin{subfigure}[t]{0.5\textwidth}
\centering
\clipexamples{playing_trumpet/7MA-1bL2-28_3_13}\\
\clipexamples{playing_trumpet/944YsAoEOS0_1_11}\\
\caption{playing trumpet}
\end{subfigure}

\begin{subfigure}[t]{0.5\textwidth}
\centering
\clipexamples{braiding_hair/a7uuFmI-D0c_235_245}\\
\clipexamples{braiding_hair/lwVPmf2Y-a8_212_222}\\
\caption{braiding hair}
\end{subfigure}%
~
\begin{subfigure}[t]{0.5\textwidth}
\centering
\clipexamples{brushing_hair/hrFrUabqA28_104_114}\\
\clipexamples{brushing_hair/Lh0ERgO65OA_196_206}\\
\caption{brushing hair}
\end{subfigure}

\begin{subfigure}[t]{0.5\textwidth}
\centering
\clipexamples{dribbling_basketball/0tb_a3LqaF4_130_140}\\
\clipexamples{dribbling_basketball/8FOCCkVPk5A_32_42}\\
\caption{dribbling basketball}
\end{subfigure}%
~
\begin{subfigure}[t]{0.5\textwidth}
\centering
\clipexamples{dunking_basketball/8zP3wA1qAF8_463_473}\\
\clipexamples{dunking_basketball/ckKXfuYL3Ag_18_28}\\
\caption{dunking basketball}
\end{subfigure}

  \caption{Example classes from the Kinetics dataset. Best seen in colour and with zoom. Note that in some cases a single image is not enough for recognizing the action (e.g.\ ``headbanging") or distinguishing classes (``dribbling basketball" vs ``dunking basketball"). The dataset contains: Singular Person Actions (e.g.\ ``robot dancing", "stretching leg"); Person-Person Actions (e.g.\ ``shaking hands", "tickling"); Person-Object Actions (e.g.\ ``riding a bike"); same verb different objects (e.g.\ ``playing violin", ``playing trumpet"); and same object different verbs (e.g.\ ``dribbling basketball", ``dunking basketball").
  These are realistic (amateur) videos -- there is often significant camera shake, for instance.}
\label{fig:samples}
\end{figure*}

\section{How the Dataset was Built \label{collection}}

In this section we describe the collection process: how candidate videos were
obtained from YouTube, and then the processing pipeline that was used to select
the candidates and clean up the dataset. We then discuss possible biases in the dataset due to the collection process.

\paragraph{Overview:} clips for each class were obtained by first searching on YouTube for
candidates, and then using Amazon Mechanical Turkers (AMT) to decide
if the clip contains the action or not. Three or more confirmations
(out of five) were required before a clip was accepted. The
dataset was de-duped, by checking that only one clip is taken from
each video, and that clips do not contain common video material. Finally, classes were 
checked for overlap and de-noised.

We now describe these stages in more detail.

\subsection{Stage 1: Obtaining an action list}

Curating a large list of human actions is challenging, as there is no
single listing available at this scale with suitable visual action classes.
Consequently, we had to combine numerous sources
together with our own observations of actions that surround us. These
sources include: (i) {\bf Action datasets} -- existing datasets like
ActivityNet~\cite{caba2015activitynet}, HMDB~\cite{Kuehne11},
UCF101~\cite{soomro2012ucf101}, MPII Human
Pose~\cite{andriluka20142d}, ACT~\cite{Wang_Transformation} have
useful classes and a suitable sub set of these were used; (ii) {\bf
Motion capture} -- there are a number of motion capture datasets which
we looked through and extracted file titles. These titles described
the motion within the file and were often quite creative; 
and, (iii)
{\bf Crowdsourced} --  we asked Mechanical Turk workers to come up with a more appropriate action if the label we had presented to them for a clip was incorrect.

\subsection{Stage 2: Obtaining candidate clips}

The chosen method and steps are detailed below which combine
a number of different internal efforts:

\paragraph{Step 1: obtaining videos.}
Videos are drawn from the YouTube corpus by matching video titles with the
Kinetics actions list.

\paragraph{Step 2: temporal positioning within a video.}
Image classifiers are available for a large number of human actions.
These classifiers are obtained by tracking user actions on Google
Image Search.  For example, for a search query ``climbing tree'', user
relevance feedback on images is collected by aggregating across the
multiple times that that search query is issued. This relevance
feedback is used to select a high-confidence set of images that can
be used to train a ``climbing tree'' image classifier.  These
classifiers are run at the frame level over the videos found in
step~1, and clips extracted around the top $k$ responses (where $k=2$).

It was found that the action list had a better match to relevant
classifiers if action verbs are formatted to end with `ing'. Thinking
back to image search, this makes sense as typically if you are
searching for an example of someone performing an action you would
issue queries like `running man' or `brushing hair' over other tenses
like `man ran' or `brush hair'.

The output of this stage is a large number of videos and a position in
all of them where one of the actions is potentially occurring.  10
second clips are created by taking 5 seconds either side of that
position (there are length exceptions when the position is within 5
seconds of the start or end of the video leading to a shorter clip
length). The clips are then passed onto the next stage of cleanup
through human labelling.

\subsection{Stage 3: Manual labelling process}

The key aim of this stage was to identify whether the supposed action
was actually occurring during a clip or not. A human was required in
the loop for this phase and we chose to use Amazon's Mechanical Turk (AMT)
for the task due to the large numbers of high quality workers using
the platform.

\begin{figure*}
  \centering
    \includegraphics[width=0.95\textwidth]{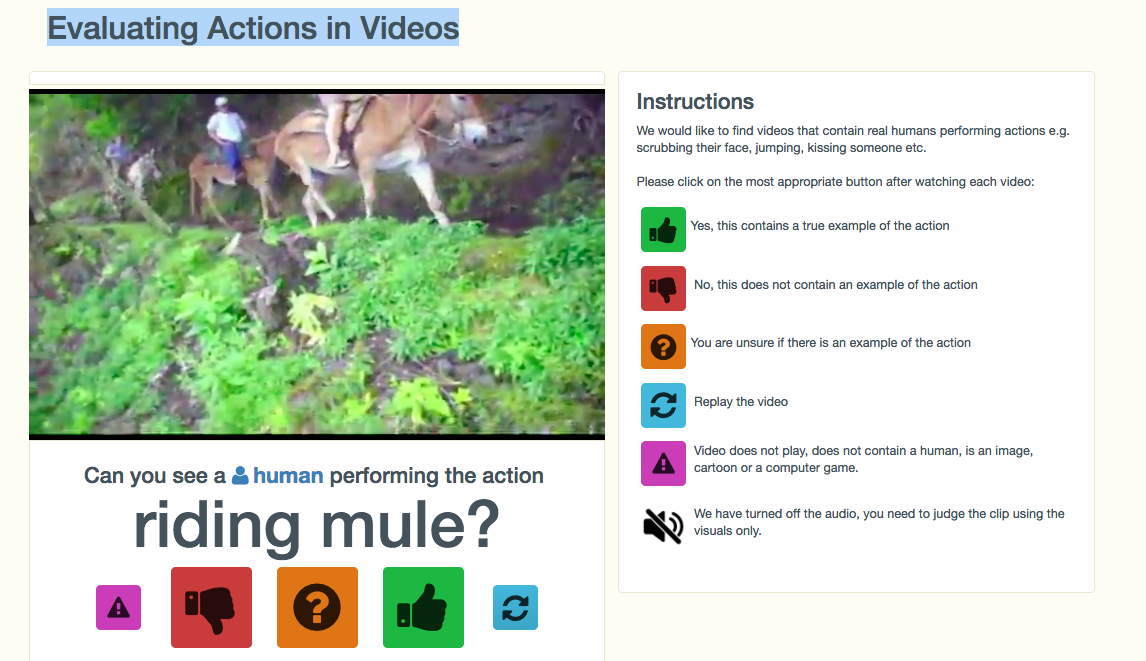}
    \caption{Labeling interface used in Mechanical Turk.}
    \label{fig:interface}
\end{figure*}

A single-page webapp was built for the labelling task and optimised to
maximise the number of clips presented to the workers whilst
maintaining a high quality of annotation. The labelling interface is
shown in figure~\ref{fig:interface}. The user interface design and theme were chosen to
differentiate the task from many others on the platform as well as
make the task as stimulating and engaging as possible. This certainly
paid off as the task was one of the highest rated on the platform and
would frequently get more than 400 distinct workers as soon as a new
run was launched.

The workers were given clear instructions at the beginning. There were
two screens of instruction, the second reinforcing the first. After
acknowledging they understood the task they were presented with a
media player and several response icons. The interface would fetch a
set of videos from the available pool for the worker at that moment
and embed the first clip. The task consisted of 20 videos each with a
different class where possible; we randomised all the videos and
classes to make it more interesting for the workers and prevent them
from becoming stuck on classes with low yields. Two of the video slots
were used by us to inject groundtruth clips. This allowed us to get an
estimate of the accuracy for each worker. If a worker fell below a
50\% success rating on these, we showed them a `low accuracy' warning
screen. This helped address many low accuracies.

In the labelling interface, workers were asked the question ``Can you
see a human performing the action {\tt class-name}?''. The following response
options were available on the interface as icons:
\begin{itemize}
\item Yes, this contains a true example of the action
\item No, this does not contain an example of the action
\item You are unsure if there is an example of the action
\item Replay the video
\item Video does not play, does not contain a human, is an image, cartoon or a computer game.
\end{itemize}
When a worker responded with `Yes' we also asked the question ``Does
the action last for the whole clip?'' in order to use this signal later
during model training.

Note, the AMT workers didn't have access to the 
audio to ensure that the video can be classified purely based on its visual content.

In order for a clip to be added to the dataset, it needed to receive
at least 3 positive responses from workers. We allowed each clip to be
annotated 5 times except if it had been annotated by more than 2 of a
specific response. For example, if 3 out of 3 workers had said it did
not contain an example of the action we would immediately remove it
from the pool and not continue until 5 workers had annotated it.

Due to the large scale of the task it was necessary to quickly remove
classes that were made up of low quality or completely irrelevant
candidates. Failing to do this would have meant that we spent a lot of money
paying workers to mark videos as negative or bad. Accuracies for each
class were calculated after 20 clips from that class had been
annotated. We adjusted the accuracy threshold between runs but would
typically start at a high accuracy of 50\% (1 in 2 videos were
expected to contain the action).

Following annotating, the video ids, clip times and labels were exported from the database and handed on to be used for model training.

\paragraph{What we learnt:}
We found that more specific classes like `riding mule' were producing
much less noise than more general classes like `riding'. However,
occasionally using more general classes was a benefit as they could
subsequently be split into a few distinct classes that were not
previously present and the candidates resent out to workers
e.g.\ `gardening' was split into `watering plants', `trimming trees'
and `planting trees'.

The amount of worker traffic that the task generated meant that we
could not rely on direct fetching and writes to the database even with
appropriate indexes and optimised queries. We therefore created many
caches which were made up of groups of clips for each worker. When a
worker started a new task, the interface would fetch a set of clips
for that specific worker. The cache was replenished often by
background processes as clips received a sufficient number of
annotations. This also negated labelling collisions where previously $>1$
worker might pick up the same video to annotate and we would quickly
exceed 5 responses for any 1 clip.

\subsection{Stage 4: Cleaning up and de-noising}

One of the dataset design goals was having a single clip from each given video sequence, different from existing datasets which slice videos containing repetitive actions into many (correlated) training examples. We also employed mechanisms for identifying structural problems as we grew the dataset, such as repeated classes due to synonymy or different word order (e.g. riding motorbike, riding motorcycle), classes that are too general and co-occur with many others (e.g. talking) and which are problematic for typical 1-of-K classification learning approaches (instead of multi-label classification). We will now describe these procedures.

\paragraph{De-duplicating videos.}

We de-duplicated videos using two complementary approaches. First, in
order to have only one clip from each YouTube link, we randomly
selected a single clip from amongst those validated by Turkers for
that video. This stage filtered out around 20\% of Turker-approved
examples, but we visually found that it still left many
duplicates. The reason is that YouTube users often create videos
reusing portions of other videos, for example as part of video
compilations or promotional adverts. Sometimes they are cropped,
resized and generally pre-processed in different ways (but, nevertheless, the image classifier could localize the same clip).  So even though
each clip is from a distinct video there were still duplications.

We devised a process for de-duplicating across YouTube links which operated independently for each class. First we computed Inception-V1 \cite{ioffe2015batch} feature vectors (taken after last average pooling layer) on $224 \times 224$ center crops of 25 uniformly sampled frames from each video, which we then averaged. Afterwards we built a class-wise matrix having all cosine similarities between these feature vectors and thresholded it. Finally, we computed connected components and kept a random example from each. We found this to work well for most classes using the same threshold of 0.97, but adjusted it in a few cases where classes were visually similar, such as some taking place in the snow or in the water. This process reduced the number of Turker-approved examples by a further 15\%.

\paragraph{Detecting noisy classes.}

Classes can be `noisy' in that they may overlap with other classes or they may contain several quite distinct (in terms of the action) groupings due to an ambiguity in the class name. For example, `skipping' can be `skipping with a rope' and also `skipping stones across water'. We trained two-stream action classifiers \cite{simonyan2014two} repeatedly throughout the dataset development to identify these noise classes. This allowed us to find the top confusions for each class, which sometimes were clear even by just verifying the class names (but went unnoticed due to the scale of the dataset), and other times required eyeballing the data to understand if the confusions were alright and the classes were just difficult to distinguish because of shortcomings of the model. We merged, split or outright removed classes based on these detected confusions.

\paragraph{Final filtering.}

After all the data was collected, de-duplicated and the classes were selected, we ran a final manual clip filtering stage. Here the class scores from the two-stream model were again useful as they allowed sorting the examples from most confident to least confident -- a measure of how prototypical they were. We found that noisy examples were often among the lowest ranked examples and focused on those. The ranking also made adjacent any remaining duplicate videos, which made it easier to filter out those too.

\subsection{Discussion: dataset bias I}

We are familiar with the notion of dataset bias leading to lack of
generalization: where a classifier trained on one dataset, e.g.\
Caltech 256~\cite{griffin2007caltech}, does not perform well when tested on another,
e.g.\ PASCAL VOC~\cite{everingham2015pascal}. Indeed it is even possible to train a classifier to
identify which dataset an image belongs to~\cite{torralba2011unbiased}.

There is another sense of bias which could arise from unbalanced
categories {\em within} a dataset. For example, gender imbalance in a
training set could lead to a corresponding performance bias for
classifiers trained on this set. There are precedents for this, e.g.\ in publicly available face detectors not being race
agnostic\footnote{\url{https://www.media.mit.edu/posts/media-lab-student-recognized-for-fighting-bias-in-machine-learning/
}}, and more recently in learning a semantic bias in written texts~\cite{Caliskan183}. It is thus an important question as to whether
Kinetics leads to such bias.

To this end we carried out a preliminary study on (i) whether the data
for each action class of Kinetics is gender balanced, and (ii) if,
there is an imbalance, whether it leads to a biased performance of the
action classifies. 

The outcome of (i) is that in 340 action classes
out of the 400, the data is either not dominated by a single gender, or it is mostly not possible to
determine the gender -- the latter arises in classes where, for
example, only hands appear, or the `actors' are too small or heavily
clothed. The classes that do show gender imbalance include `shaving beard' and `dunking basketball', that are mostly male, and
 'filling eyebrows' and `cheerleading', that are mostly female.

The outcome of (ii) for these classes we found little evidence of
classifier bias for action classes with gender imbalance. For example
in `playing poker', which tends to have more male players, all videos with female players are correctly
classified. The same happens for `Hammer throw'. We can conjecture that this lack of bias is because the
classifier is able to make use of both the objects involved in an
action as well as the motion patterns, rather than simply physical appearance.

Imbalance can also be examined on other `axes', for example age and
race.  Again, in a preliminary investigation we found very little
clear bias. There is one exception where there is clear bias to babies
-- in `crying', where many of the videos of non-babies crying are
misclassified; another example is `wrestling', where the opposite happens: adults wrestling in a ring seem to be better classified 
than children wrestling in their homes, but it is hard to tell whether the deciding factor is age or the scenes where the actions happen.  Nevertheless, these issues of
dataset imbalance and any resulting classifier bias warrant a more thorough investigation, and we return to this in
section~\ref{sec:conclusion}.

\subsection{Discussion: dataset bias II }

Another type of bias could arise because classifiers are involved in
the dataset collection pipeline: it could be that these
classifiers lead to a reduction in the visual variety of the clips
obtained, which in turn leads to a bias in the action classifier
trained on these clips.  In more detail, although the videos are
selected based on their title (which is provided by the person
uploading the video to YouTube), the {\em position} of the candidate
clip within the video is provided by an image (RGB) classifier, as
described above.  In practice, using a classifier at this point does
not seem to constrain the variety of the clips -- since the video is
about the action, the particular frame chosen as part of the clip may
not be crucial; and, in any case, the clip contains hundreds of more
frames where the appearance (RGB) and motion can vary considerably.
For these reasons we are not so concerned about the intermediate use
of image classifiers.

\section{Benchmark Performance}
\label{baselines}

In this section we first briefly describe three standard ConvNet
architectures for human action recognition in video. We then use these
architectures as baselines and compare their performance by training
and testing on the Kinetics dataset. We also include their performance
on UCF-101 and HMDB-51.

\begin{figure*}
  \centering
    \includegraphics[width=0.95\textwidth]{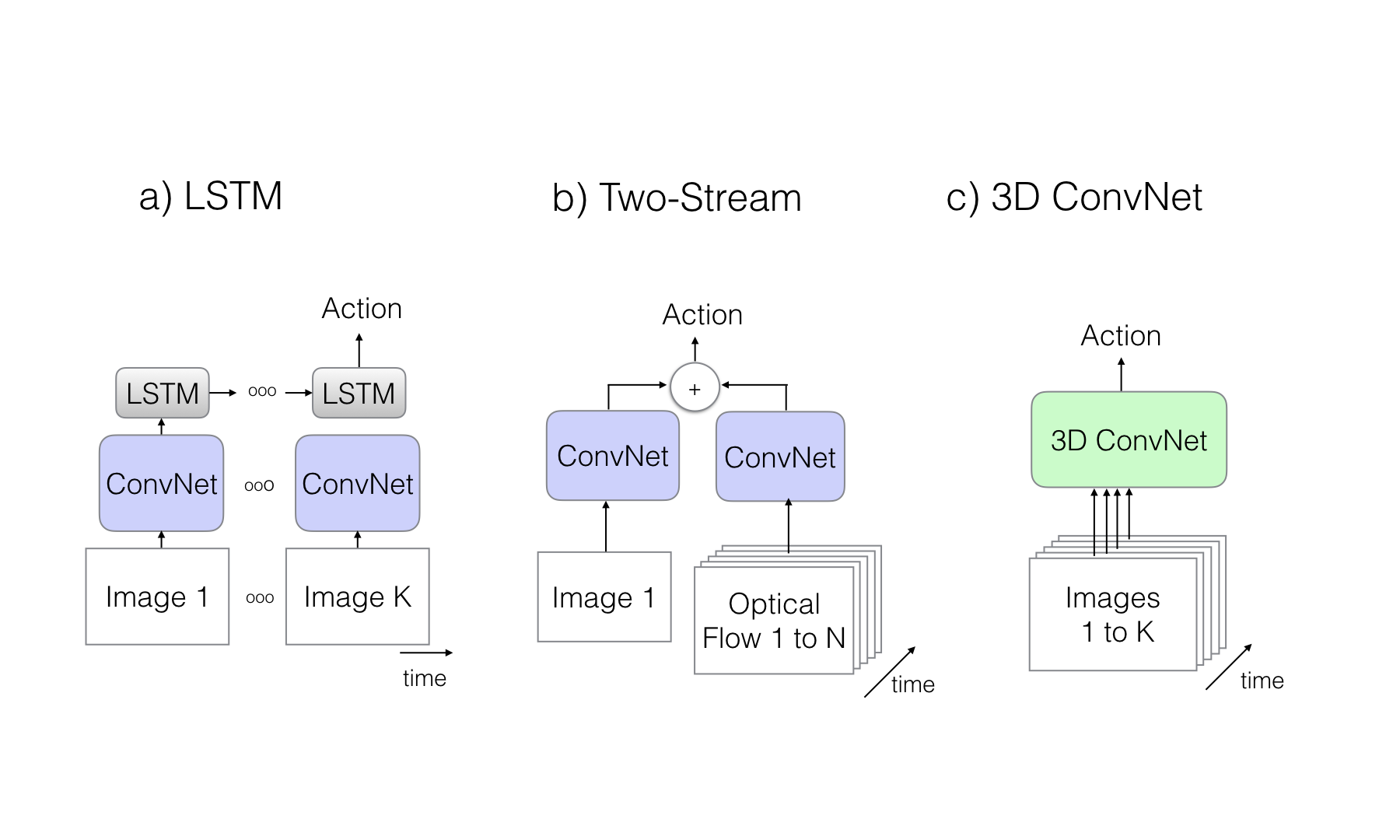}
    \caption{Video architectures used as baseline human action classifiers.}
    \label{fig:architectures}
\end{figure*}

We consider three typical approaches for video classification:
 ConvNets with an LSTM on top~\cite{yue2015beyond,donahue2015long};
two-stream
networks~\cite{simonyan2014two,feichtenhofer2016convolutional}; and a
3D ConvNet~\cite{taylor2010convolutional,ji20133d,tran2015learning}. There
have been many improvements over these basic architectures,
e.g.~\cite{feichtenhofer2016convolutional}, but our intention
here is not to perform a thorough study on what is the very best architecture on Kinetics, but instead to provide an indication of the level of difficulty of the dataset.
A rough graphical overview of the three types of architectures we
compare is shown in figure~\ref{fig:architectures}, and the
specification of their temporal interfaces is given in 
table~\ref{tab:temporal_interfaces}.

For the experiments on the Kinetics dataset all three architectures are trained from 
scratch using Kinetics. However, for 
the experiments on UCF-101
and HMDB-51 the architectures (apart from the 3D ConvNet) are pre-trained
on ImageNet (since these datasets are too small to train the architectures from scratch).

\subsection{ConvNet+LSTM}

The high performance of image classification networks makes it appealing to try to reuse them  with as minimal change as possible for video. This can be achieved by using them to extract features independently from each frame then pooling their predictions across the whole video \cite{karpathy2014large}. This is in the spirit of bag of words image modeling approaches \cite{wangICCV13,niebles2008unsupervised,laptev2008learning}, but while convenient in practice, it has the issue of entirely ignoring temporal structure (e.g.\ models can't potentially distinguish opening from closing a door). 

In theory, a more satisfying approach is to add a recurrent layer to the model \cite{yue2015beyond,donahue2015long}, such as an LSTM, which can encode state, and capture temporal ordering and long range dependencies. We position an LSTM layer with batch normalization (as proposed by Cooijmans {\it et al.}~\cite{cooijmans2016recurrent}) after the last average pooling layer of a ResNet-50 model \cite{he2016deep}, with 512 hidden units. We then add a fully connected layer on top of the output of the LSTM for the multi-way classification. At test time the classification is taken from the model output for the last frame. 

\subsection{Two-Stream networks}

LSTMs on features from the last layers of ConvNets can model high-level variation, but may not be able to capture fine low-level motion which is critical in many cases. It is also expensive to train as it requires unrolling the network through multiple frames for backpropagation-through-time. 

A different, very practical approach, introduced by Simonyan and Zisserman~\cite{simonyan2014two}, models short temporal snapshots of videos by averaging the predictions from a single RGB frame and a stack of $10$ externally computed optical flow frames, after passing them through two replicas of an ImageNet-pretrained ConvNet. The flow stream has an adapted input convolutional layer with twice as many input channels as flow frames (because flow has two channels, horizontal and vertical), and at test time multiple snapshots are sampled from the video and the action prediction is averaged. This was shown to get very high performance on existing benchmarks, while being very efficient to train and test.

\subsection{3D ConvNets}

3D ConvNets~\cite{taylor2010convolutional,ji20133d,tran2015learning} 
seem like a natural approach to video modeling. They are
just like standard 2D convolutional networks, but with spatio-temporal
filters, and have a
very interesting characteristic: they directly create hierarchical
representations of spatio-temporal data. One issue with these models
is that they have many more parameters than 2D ConvNets because of the
additional kernel dimension, and this makes them harder to train.
Also, they seem to preclude the benefits of ImageNet pre-training and
previous work has defined relatively shallow custom architectures and
trained them from scratch
\cite{taylor2010convolutional,ji20133d,tran2015learning,karpathy2014large}. Results
on benchmarks have shown promise but have not yet matched the
state-of-the-art, possibly because they require more training data than
their 2D counterparts. Thus 3D ConvNets are a good candidate for
evaluation on our larger dataset.

For this paper we implemented a small variation of C3D \cite{tran2015learning}, which has $8$ convolutional layers, $5$ pooling layers and $2$ fully connected layers at the top. The inputs to the model are short $16$-frame clips with $112 \times 112$-pixel crops. Differently from the original paper we use batch normalization after all convolutional and fully connected layers. Another difference to the original model is in the first pooling layer, where we use a temporal stride of $2$ instead of $1$, which reduces the memory footprint and allows for bigger batches -- this was important for batch normalization (especially after the fully connected layers, where there is no weight tying). Using this stride we were able to train with 15 videos per batch per GPU using standard K40 GPUs.

At test time, we split the video uniformly into crops of 16 frames and apply the classifier separately on each. We then average the class scores, as in the original paper.

\begin{table*}[t]
\centering
\begin{tabular}{| c | c | c | c | c | c| }
  \hline
  
  \multirow{2}{*}{\parbox{1.5cm}{Method}} & \multirow{2}{*}{\parbox{1.5cm}{\#Params}} &
  \multicolumn{2}{|c|}{Training} &
  \multicolumn{2}{|c|}{Testing} \\ \cline{3-6}

   & & \# Input Frames & Temporal Footprint & \# Input Frames & Temporal Footprint  \\ \hline \hline
   
(a) ConvNet+LSTM & 29M & 25 rgb  &  5s &  50 rgb  & 10s \\ \hline
(b) Two-Stream  & 48M & 1 rgb, 10 flow  & 0.4s  & 25 rgb, 250 flow & 10s \\ \hline
(c) 3D-ConvNet & 79M & 16 rgb & 0.64s & 240 rgb & 9.6s \\ \hline
\end{tabular} 
\vspace{5pt}
\caption{Number of parameters and temporal input sizes of the models. ConvNet+LSTM and Two-Stream use ResNet-50 ConvNet modules.} 
\label{tab:temporal_interfaces}
\end{table*}

\begin{table*}[t]
\begin{center}
    \begin{tabular}{| l | c | c | c || c | c | c || c | c | c |}
    \hline
 &  \multicolumn{3}{|c||}{UCF-101} & \multicolumn{3}{c||}{HMDB-51}  & \multicolumn{3}{|c|}{Kinetics} \\ \cline{2-4} \cline{5-7} \cline{8-10}
Architecture            & RGB   & Flow  & RGB+Flow &  RGB   & Flow  & RGB+Flow & RGB   & Flow  & RGB+Flow    \\ \hline \hline
    (a) ConvNet+LSTM           & 84.3 & -- & -- & 43.9 & --  & -- & 57.0 / 79.0 & -- & --      \\ \hline
    (b) Two-Stream          & 84.2  & 85.9 & 92.5  & 51.0 & 56.9 & 63.7 & 56.0 / 77.3 & 49.5 / 71.9 & 61.0 / 81.3   \\\hline
    (c) 3D-ConvNet           & 51.6 & -- & -- & 24.3 & --  & -- & 56.1 / 79.5 & -- & --           \\ \hline
    \hline
    \end{tabular}

\end{center}
\caption{Baseline comparisons across datasets: (left) training and testing on split 1 of UCF-101; 
(middle) training and testing on split 1 of HMDB-51; (right) training and testing on Kinetics (showing top-1/top-5 performance). ConvNet+LSTM and Two-Stream use ResNet-50 ConvNet modules, pretrained on ImageNet for UCF-101 and HMDB-51 examples but not for the Kinetics experiments. Note that the Two-Stream architecture numbers on individual RGB and Flow streams can be interpreted as a simple baseline which applies a ConvNet  independently on 25 uniformly sampled frames then averages the predictions.}
\label{fig:archUCF101}
\label{fig:archKinetics}
\end{table*}

\subsection{Implementation details}

The ConvNet+LSTM and Two-Stream architecures use ResNet-50 as the base architecture. 
In the case of the Two-Stream architecture, a separate ResNet-50 is trained
independently for each stream. As noted earlier, for these architectures
the ResNet-50 model is pre-trained on ImageNet for the 
experiments on UCF-101 and HMDB-51, and trained from scratch for
experiments on Kinetics. The 3D-ConvNet is not pre-trained.

We trained the models on videos using standard SGD with momentum in
all cases, with synchronous parallelization across 64 GPUs for all
models. We trained models on Kinetics for up to 100k steps, with a 10x
reduction of learning rate when validation loss saturated, and tuned
weight decay and learning rate hyperparameters on the validation set
of Kinetics. All the models were implemented in TensorFlow
\cite{abadi2016tensorflow}.

The original clips have variable resolution and
frame rate. In our experiments they are all normalized so that the larger image side is 340 pixels wide for models using ResNet-50 and 128 pixels wide for the 3D ConvNet. We also resample the videos so they have 25 frames per second.

Data augmentation is known to be of crucial importance for the
performance of deep architectures. We used random cropping both
spatially -- randomly cropping a $299 \times 299$ patch  (respectively $112 \times 112$ for the 3D ConvNet) -- and temporally, when
picking the starting frame among those early enough to guarantee a
desired number of frames. For shorter videos, we looped the video
 as many times as necessary to satisfy each model's input
interface. We also applied random left-right flipping consistently for
each video during training. 

At test time, we sample from up to 10 seconds of video, again looping if necessary. Better performance could be obtained by
also considering left-right flipped videos at test time and by adding
additional augmentation, such as photometric, during training. We
leave this to future work.

\subsection{Baseline evaluations \label{comp_arch}}

In this section we compare the performance of the three baseline architectures whilst varying the dataset used for training and testing.

Table~\ref{fig:archUCF101} shows the 
classification accuracy when training and testing on either UCF-101, HMDB-51 or Kinetics. We train and test on split~1 of UCF-101 and HMDB-51, and on the train/val set and held-out test set of Kinetics. 

There are several noteworthy observations. First, the performance is
far lower on Kinetics than on UCF-101, an indication of the different
levels of difficulty of the two datasets. On the other hand, the
performance on HMDB-51 is worse than on Kinetics -- it seems to have a
truly difficult test set, and it was designed to be difficult to
appearance-centered methods, while having little training data. The
parameter-rich 3D-ConvNet model is not pre-trained on ImageNet, unlike
the other baselines. This translates into poor performance on all
datasets but especially on UCF-101 and HMDB-51 -- on Kinetics it is
much closer to the performance of the other models,
thanks to the much larger training set of Kinetics.

\begin{figure}
  \centering
    \includegraphics[width=0.45\textwidth]{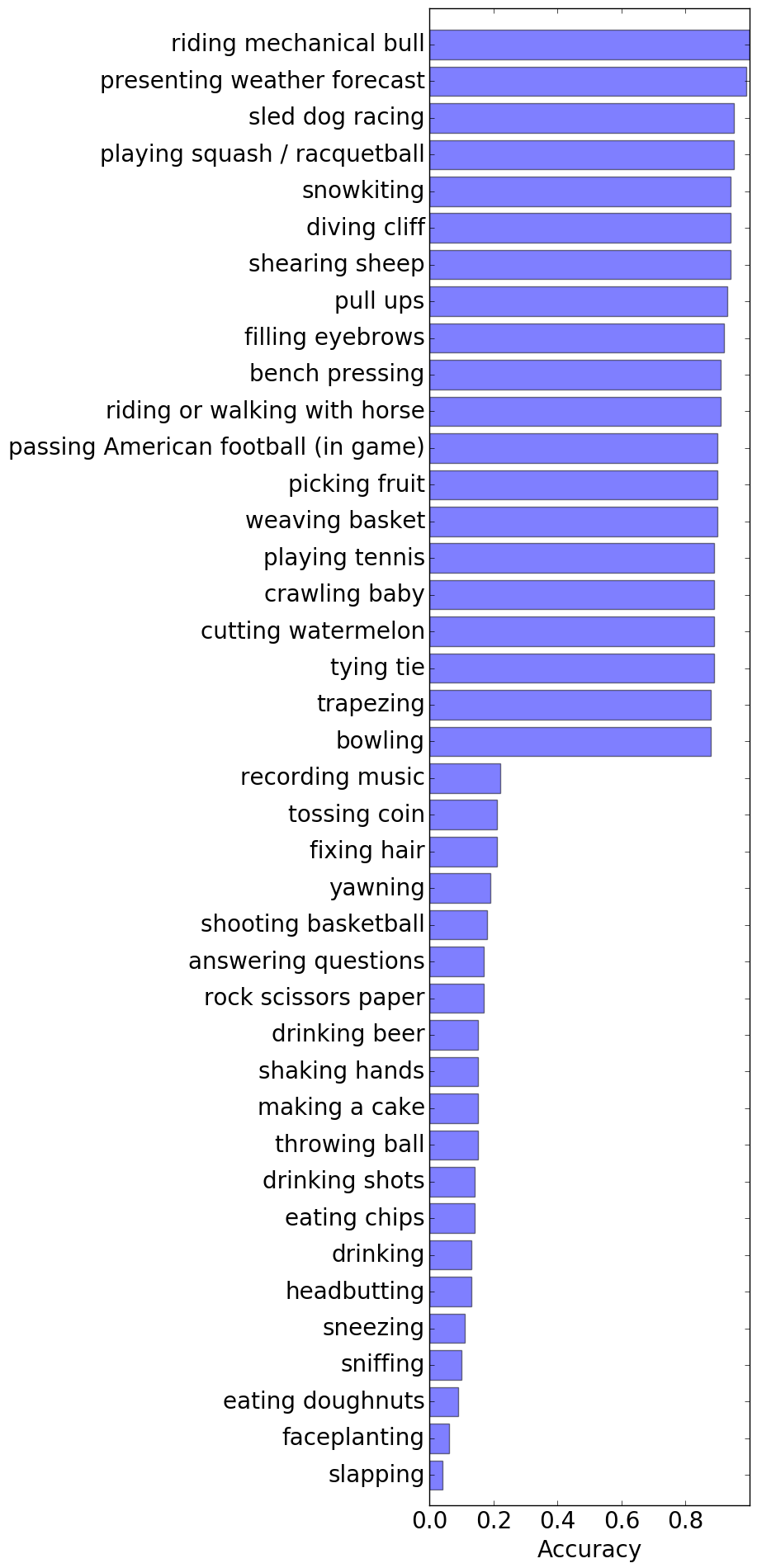}\label{fig:f1}
    \caption{List of 20 easiest and 20 hardest Kinetics classes sorted by class accuracies obtained using the two-stream model.}
    \label{fig:accuracies}
\end{figure}

\begin{itemize}
\item \textbf{Class difficulty.} We include a full list of Kinetics classes sorted by classification accuracy under the two-stream model in figure~\ref{fig:accuracies}. Eating classes are among the hardest, as they sometimes require distinguishing what is being eaten, such as hotdogs, chips and doughnuts -- and these may appear small and already partially consumed, in the video. Dancing classes are also hard, as well as classes centered on a specific body part, such as ``massaging feet", or ``shaking  head".

\item \textbf{Class confusion}. The top 10 class confusions are provided in table~\ref{tab:confusion}. They mostly correspond to fine-grained distinctions that one would expect to be hard, for example `long jump' and `triple jump', confusing burger with doughnuts. The confusion between `swing dancing' and `salsa dancing' raises the question of how accurate motion modeling is in the two-stream model, since `swing dancing' is typically much faster-paced and has a peculiar style that makes it easy for humans to distinguish from salsa.

\item \textbf{Classes where motion matters most.} 
We tried to analyze for which classes motion is more
important and which ones were recognized correctly using just
appearance information, by comparing the recognition accuracy ratios
when using the flow and RGB streams of the  two-stream model in isolation. We show
the five classes where this ratio is largest and smallest in
table~\ref{tab:motion_appearance}.

\end{itemize}

\begin{table*}[ht]
\centering
\begin{tabular}{| c | c | c | }
  \hline \textbf{Class 1} & \textbf{Class 2} & \textbf{confusion} \\
  \hline\hline
`riding mule' & `riding or walking with horse' & 40\% \\
\hline 
`hockey stop' & `ice skating' & 36\% \\
\hline 
`swing dancing' & `salsa dancing' & 36\% \\
\hline 
`strumming guitar' & `playing guitar' & 35\% \\
\hline 
`shooting basketball' & `playing basketball' & 32\% \\
\hline 
`cooking sausages' & `cooking chicken' & 29\% \\
\hline 
`sweeping floor' & `mopping floor' & 27\% \\
\hline 
`triple jump' & `long jump' & 26\% \\
\hline 
`doing aerobics' & `zumba' & 26\% \\
\hline 
`petting animal (not cat)' & `feeding goats' & 25\% \\
\hline 
`shaving legs' & `waxing legs' & 25\% \\
\hline 
`snowboarding' & `skiing (not slalom or crosscountry)' & 22\% \\
\hline 
\end{tabular}
\vspace{5pt}
\caption{Top-12 class confusions in Kinetics, using the two-stream model.}
\label{tab:confusion}
\end{table*}

\begin{table}[ht]
\centering
\begin{tabular}{| c | c | c | }
  \hline
  \textbf{Class} & \textbf{Flow/RGB accuracy ratio} \\ \hline\hline 

`rock scissors paper' & 5.3 \\
\hline 
`sword fighting' & 3.1 \\
\hline 
`robot dancing' & 3.1 \\
\hline 
`air drumming' & 2.8 \\
\hline 
`exercising arm' & 2.5 \\
\hline 
`making a cake' & 0.1  \\
\hline 
`cooking sausages' & 0.1 \\
\hline 
`sniffing' & 0.1  \\
\hline 
`eating cake' & 0.0  \\
\hline 
`making a sandwich' & 0.0 \\
\hline 
\hline 
\end{tabular}
\vspace{5pt}
\caption{Classes with largest and smallest ratios of recognition accuracy when using flow and RGB. The highest ratios correspond to when flow does better, the smallest to when RGB does better. We also evaluated the ratios of rgb+flow to rgb accuracies and the ordering was quite similar.}
\label{tab:motion_appearance}
\end{table}

\section{Conclusion}
\label{sec:conclusion}

We have described the Kinetics Human Action Video dataset, which has
an order of magnitude more videos than previous datasets of its type.
We have also discussed the procedures we employed collecting
the data and for ensuring its quality. We have shown that the
performance of standard existing models on this dataset is much lower
than on UCF-101 and on par with HMDB-51, whilst allowing large
models such as 3D ConvNets to be trained from scratch, unlike the existing human
action datasets.

We have also carried out a preliminary analysis of dataset imbalance and whether this leads to bias in the classifiers trained on the dataset. We found little evidence that the resulting classifiers demonstrate bias along sensitive axes, such as across gender. This is however a complex area that deserves further attention. We leave a thorough analysis for future work, in collaboration with specialists from complementary areas, namely social scientists and critical humanists.

We will release trained baseline models (in TensorFlow), so that they can be used, for example, to generate features for new action classes.

\subsection*{Acknowledgements:} The collection of this dataset was funded by DeepMind. We are very grateful for help from Andreas Kirsch, John-Paul Holt, Danielle Breen, Jonathan Fildes, James Besley and Brian Carver. We are grateful for advice and comments from Tom Duerig, Juan Carlos Niebles, Simon Osindero, Chuck Rosenberg and Sean Legassick; we would also like to thank  Sandra and Aditya for data clean up.

{\small
\bibliographystyle{ieee}
\bibliography{references}
}

\newpage


\appendix

\section{List of Kinetics Human Action Classes}
This is the list of classes included in the human action video
dataset. The number of clips for each action class is given by the
number in brackets following each class name.

\begin{enumerate}
\itemsep0em 
\item abseiling (1146)
\item air drumming (1132)
\item answering questions (478)
\item applauding (411)
\item applying cream (478)
\item archery (1147)
\item arm wrestling (1123)
\item arranging flowers (583)
\item assembling computer (542)
\item auctioning (478)
\item baby waking up (611)
\item baking cookies (927)
\item balloon blowing (826)
\item bandaging (569)
\item barbequing (1070)
\item bartending (601)
\item beatboxing (943)
\item bee keeping (430)
\item belly dancing (1115)
\item bench pressing (1106)
\item bending back (635)
\item bending metal (410)
\item biking through snow (1052)
\item blasting sand (713)
\item blowing glass (1145)
\item blowing leaves (405)
\item blowing nose (597)
\item blowing out candles (1150)
\item bobsledding (605)
\item bookbinding (914)
\item bouncing on trampoline (690)
\item bowling (1079)
\item braiding hair (780)
\item breading or breadcrumbing (454)
\item breakdancing (948)
\item brush painting (532)
\item brushing hair (934)
\item brushing teeth (1149)
\item building cabinet (431)
\item building shed (427)
\item bungee jumping (1056)
\item busking (851)
\item canoeing or kayaking (1146)
\item capoeira (1092)
\item carrying baby (558)
\item cartwheeling (616)
\item carving pumpkin (711)
\item catching fish (671)
\item catching or throwing baseball (756)
\item catching or throwing frisbee (1060)
\item catching or throwing softball (842)
\item celebrating (751)
\item changing oil (714)
\item changing wheel (459)
\item checking tires (555)
\item cheerleading (1145)
\item chopping wood (916)
\item clapping (491)
\item clay pottery making (513)
\item clean and jerk (902)
\item cleaning floor (874)
\item cleaning gutters (598)
\item cleaning pool (447)
\item cleaning shoes (706)
\item cleaning toilet (576)
\item cleaning windows (695)
\item climbing a rope (413)
\item climbing ladder (662)
\item climbing tree (1120)
\item contact juggling (1135)
\item cooking chicken (1000)
\item cooking egg (618)
\item cooking on campfire (403)
\item cooking sausages (467)
\item counting money (674)
\item country line dancing (1015)
\item cracking neck (449)
\item crawling baby (1150)
\item crossing river (951)
\item crying (1037)
\item curling hair (855)
\item cutting nails (560)
\item cutting pineapple (712)
\item cutting watermelon (767)
\item dancing ballet (1144)
\item dancing charleston (721)
\item dancing gangnam style (836)
\item dancing macarena (958)
\item deadlifting (805)
\item decorating the christmas tree (612)
\item digging (404)
\item dining (671)
\item disc golfing (565)
\item diving cliff (1075)
\item dodgeball (595)
\item doing aerobics (461)
\item doing laundry (461)
\item doing nails (949)
\item drawing (445)
\item dribbling basketball (923)
\item drinking (599)
\item drinking beer (575)
\item drinking shots (403)
\item driving car (1118)
\item driving tractor (922)
\item drop kicking (716)
\item drumming fingers (409)
\item dunking basketball (1105)
\item dying hair (1072)
\item eating burger (864)
\item eating cake (494)
\item eating carrots (516)
\item eating chips (749)
\item eating doughnuts (528)
\item eating hotdog (570)
\item eating ice cream (927)
\item eating spaghetti (1145)
\item eating watermelon (550)
\item egg hunting (500)
\item exercising arm (416)
\item exercising with an exercise ball (438)
\item extinguishing fire (602)
\item faceplanting (441)
\item feeding birds (1150)
\item feeding fish (973)
\item feeding goats (1027)
\item filling eyebrows (1085)
\item finger snapping (825)
\item fixing hair (676)
\item flipping pancake (720)
\item flying kite (1063)
\item folding clothes (695)
\item folding napkins (874)
\item folding paper (940)
\item front raises (962)
\item frying vegetables (608)
\item garbage collecting (441)
\item gargling (430)
\item getting a haircut (658)
\item getting a tattoo (737)
\item giving or receiving award (953)
\item golf chipping (699)
\item golf driving (836)
\item golf putting (1081)
\item grinding meat (415)
\item grooming dog (613)
\item grooming horse (645)
\item gymnastics tumbling (1143)
\item hammer throw (1148)
\item headbanging (1090)
\item headbutting (640)
\item high jump (954)
\item high kick (825)
\item hitting baseball (1071)
\item hockey stop (468)
\item holding snake (430)
\item hopscotch (726)
\item hoverboarding (564)
\item hugging (517)
\item hula hooping (1129)
\item hurdling (622)
\item hurling (sport) (836)
\item ice climbing (845)
\item ice fishing (555)
\item ice skating (1140)
\item ironing (535)
\item javelin throw (912)
\item jetskiing (1140)
\item jogging (417)
\item juggling balls (923)
\item juggling fire (668)
\item juggling soccer ball (484)
\item jumping into pool (1133)
\item jumpstyle dancing (662)
\item kicking field goal (833)
\item kicking soccer ball (544)
\item kissing (733)
\item kitesurfing (794)
\item knitting (691)
\item krumping (657)
\item laughing (926)
\item laying bricks (432)
\item long jump (831)
\item lunge (759)
\item making a cake (463)
\item making a sandwich (440)
\item making bed (679)
\item making jewelry (658)
\item making pizza (1147)
\item making snowman (756)
\item making sushi (434)
\item making tea (426)
\item marching (1146)
\item massaging back (1113)
\item massaging feet (478)
\item massaging legs (592)
\item massaging person\ (672)
\item milking cow (980)
\item mopping floor (606)
\item motorcycling (1142)
\item moving furniture (426)
\item mowing lawn (1147)
\item news anchoring (420)
\item opening bottle (732)
\item opening present (866)
\item paragliding (800)
\item parasailing (762)
\item parkour (504)
\item passing American football (in game) (863)
\item passing American football (not in game) (1045)
\item peeling apples (592)
\item peeling potatoes (457)
\item petting animal (not cat) (757)
\item petting cat (756)
\item picking fruit (793)
\item planting trees (557)
\item plastering (428)
\item playing accordion (925)
\item playing badminton (944)
\item playing bagpipes (838)
\item playing basketball (1144)
\item playing bass guitar (1135)
\item playing cards (737)
\item playing cello (1081)
\item playing chess (850)
\item playing clarinet (1022)
\item playing controller (524)
\item playing cricket (949)
\item playing cymbals (636)
\item playing didgeridoo (787)
\item playing drums (908)
\item playing flute (475)
\item playing guitar (1135)
\item playing harmonica (1006)
\item playing harp (1149)
\item playing ice hockey (917)
\item playing keyboard (715)
\item playing kickball (468)
\item playing monopoly (731)
\item playing organ (672)
\item playing paintball (1140)
\item playing piano (691)
\item playing poker (1134)
\item playing recorder (1148)
\item playing saxophone (916)
\item playing squash or racquetball (980)
\item playing tennis (1144)
\item playing trombone (1149)
\item playing trumpet (989)
\item playing ukulele (1146)
\item playing violin (1142)
\item playing volleyball (804)
\item playing xylophone (746)
\item pole vault (984)
\item presenting weather forecast (1050)
\item pull ups (1121)
\item pumping fist (1009)
\item pumping gas (544)
\item punching bag (1150)
\item punching person (boxing) (483)
\item push up (614)
\item pushing car (1069)
\item pushing cart (1150)
\item pushing wheelchair (465)
\item reading book (1148)
\item reading newspaper (424)
\item recording music (415)
\item riding a bike (476)
\item riding camel (716)
\item riding elephant (1104)
\item riding mechanical bull (698)
\item riding mountain bike (495)
\item riding mule (476)
\item riding or walking with horse (1131)
\item riding scooter (674)
\item riding unicycle (864)
\item ripping paper (605)
\item robot dancing (893)
\item rock climbing (1144)
\item rock scissors paper (424)
\item roller skating (960)
\item running on treadmill (428)
\item sailing (867)
\item salsa dancing (1148)
\item sanding floor (574)
\item scrambling eggs (816)
\item scuba diving (968)
\item setting table (478)
\item shaking hands (640)
\item shaking head (885)
\item sharpening knives (424)
\item sharpening pencil (752)
\item shaving head (971)
\item shaving legs (509)
\item shearing sheep (988)
\item shining shoes (615)
\item shooting basketball (595)
\item shooting goal (soccer) (444)
\item shot put (987)
\item shoveling snow (879)
\item shredding paper (403)
\item shuffling cards (828)
\item side kick (991)
\item sign language interpreting (446)
\item singing (1147)
\item situp (817)
\item skateboarding (1139)
\item ski jumping (1051)
\item skiing (not slalom or crosscountry) (1140)
\item skiing crosscountry (477)
\item skiing slalom (539)
\item skipping rope (488)
\item skydiving (505)
\item slacklining (790)
\item slapping (465)
\item sled dog racing (775)
\item smoking (1105)
\item smoking hookah (857)
\item snatch weight lifting (943)
\item sneezing (505)
\item sniffing (399)
\item snorkeling (1012)
\item snowboarding (937)
\item snowkiting (1145)
\item snowmobiling (601)
\item somersaulting (993)
\item spinning poi (1134)
\item spray painting (908)
\item spraying (470)
\item springboard diving (406)
\item squat (1148)
\item sticking tongue out (770)
\item stomping grapes (444)
\item stretching arm (718)
\item stretching leg (829)
\item strumming guitar (472)
\item surfing crowd (876)
\item surfing water (751)
\item sweeping floor (604)
\item swimming backstroke (1077)
\item swimming breast stroke (833)
\item swimming butterfly stroke (678)
\item swing dancing (512)
\item swinging legs (409)
\item swinging on something (482)
\item sword fighting (473)
\item tai chi (1070)
\item taking a shower (378)
\item tango dancing (1114)
\item tap dancing (947)
\item tapping guitar (815)
\item tapping pen (703)
\item tasting beer (588)
\item tasting food (613)
\item testifying (497)
\item texting (704)
\item throwing axe (816)
\item throwing ball (634)
\item throwing discus (1104)
\item tickling (610)
\item tobogganing (1147)
\item tossing coin (461)
\item tossing salad (463)
\item training dog (481)
\item trapezing (786)
\item trimming or shaving beard (981)
\item trimming trees (665)
\item triple jump (784)
\item tying bow tie (387)
\item tying knot (not on a tie) (844)
\item tying tie (673)
\item unboxing (858)
\item unloading truck (406)
\item using computer (937)
\item using remote controller (not gaming) (549)
\item using segway (387)
\item vault (562)
\item waiting in line (430)
\item walking the dog (1145)
\item washing dishes (1048)
\item washing feet (862)
\item washing hair (423)
\item washing hands (916)
\item water skiing (763)
\item water sliding (420)
\item watering plants (680)
\item waxing back (537)
\item waxing chest (760)
\item waxing eyebrows (720)
\item waxing legs (948)
\item weaving basket (743)
\item welding (759)
\item whistling (416)
\item windsurfing (1114)
\item wrapping present (861)
\item wrestling (488)
\item writing (735)
\item yawning (398)
\item yoga (1140)
\item zumba (1093)
\end{enumerate}

\section{List of Parent-Child Groupings}
These lists are not exclusive and are not intended to be comprehensive. Rather, they are a guide for related human action classes.\\

\noindent \textbf{arts and crafts (12)}\\
arranging flowers\\
blowing glass\\
brush painting\\
carving pumpkin\\
clay pottery making\\
decorating the christmas tree\\
drawing\\
getting a tattoo\\
knitting\\
making jewelry\\
spray painting\\
weaving basket\\

\noindent \textbf{athletics -- jumping (6)}\\
high jump\\
hurdling\\
long jump\\
parkour\\
pole vault\\
triple jump\\

\noindent \textbf{athletics -- throwing + launching (9)}\\
archery\\
catching or throwing frisbee\\
disc golfing\\
hammer throw\\
javelin throw\\
shot put\\
throwing axe\\
throwing ball\\
throwing discus\\

\noindent \textbf{auto maintenance (4)}\\
changing oil\\
changing wheel\\
checking tires\\
pumping gas\\

\noindent \textbf{ball sports (25)}\\
bowling\\
catching or throwing baseball\\
catching or throwing softball\\
dodgeball\\
dribbling basketball\\
dunking basketball\\
golf chipping\\
golf driving\\
golf putting\\
hitting baseball\\
hurling (sport)\\
juggling soccer ball\\
kicking field goal\\
kicking soccer ball\\
passing American football (in game)\\
passing American football (not in game)\\
playing basketball\\
playing cricket\\
playing kickball\\
playing squash or racquetball\\
playing tennis\\
playing volleyball\\
shooting basketball\\
shooting goal (soccer)\\
shot put\\

\noindent \textbf{body motions (16)}\\
air drumming\\
applauding\\
baby waking up\\
bending back\\
clapping\\
cracking neck\\
drumming fingers\\
finger snapping\\
headbanging\\
headbutting\\
pumping fist\\
shaking head\\
stretching arm\\
stretching leg\\
swinging legs\\

\noindent \textbf{cleaning (13)}\\
cleaning floor\\
cleaning gutters\\
cleaning pool\\
cleaning shoes\\
cleaning toilet\\
cleaning windows\\
doing laundry\\
making bed\\
mopping floor\\
setting table\\
shining shoes\\
sweeping floor\\
washing dishes\\

\noindent \textbf{cloths (8)}\\
bandaging\\
doing laundry\\
folding clothes\\
folding napkins\\
ironing\\
making bed\\
tying bow tie\\
tying knot (not on a tie)\\
tying tie\\

\noindent \textbf{communication (11)}\\
answering questions\\
auctioning\\
bartending\\
celebrating\\
crying\\
giving or receiving award\\
laughing\\
news anchoring\\
presenting weather forecast\\
sign language interpreting\\
testifying\\

\noindent \textbf{cooking (22)}\\
baking cookies\\
barbequing\\
breading or breadcrumbing\\
cooking chicken\\
cooking egg\\
cooking on campfire\\
cooking sausages\\
cutting pineapple\\
cutting watermelon\\
flipping pancake\\
frying vegetables\\
grinding meat\\
making a cake\\
making a sandwich\\
making pizza\\
making sushi\\
making tea\\
peeling apples\\
peeling potatoes\\
picking fruit\\
scrambling eggs\\
tossing salad\\

\noindent \textbf{dancing (18)}\\
belly dancing\\
breakdancing\\
capoeira\\
cheerleading\\
country line dancing\\
dancing ballet\\
dancing charleston\\
dancing gangnam style\\
dancing macarena\\
jumpstyle dancing\\
krumping\\
marching\\
robot dancing\\
salsa dancing\\
swing dancing\\
tango dancing\\
tap dancing\\
zumba\\

\noindent \textbf{eating + drinking (17)}\\
bartending\\
dining\\
drinking\\
drinking beer\\
drinking shots\\
eating burger\\
eating cake\\
eating carrots\\
eating chips\\
eating doughnuts\\
eating hotdog\\
eating ice cream\\
eating spaghetti\\
eating watermelon\\
opening bottle\\
tasting beer\\
tasting food\\

\noindent \textbf{electronics (5)}\\
assembling computer\\
playing controller\\
texting\\
using computer\\
using remote controller (not gaming)\\

\noindent \textbf{garden + plants (10)}\\
blowing leaves\\
carving pumpkin\\
chopping wood\\
climbing tree\\
decorating the christmas tree\\
egg hunting\\
mowing lawn\\
planting trees\\
trimming trees\\
watering plants\\

\noindent \textbf{golf (3)}\\
golf chipping\\
golf driving\\
golf putting\\

\noindent \textbf{gymnastics (5)}\\
bouncing on trampoline\\
cartwheeling\\
gymnastics tumbling\\
somersaulting\\
vault\\

\noindent \textbf{hair (14)}\\
braiding hair\\
brushing hair\\
curling hair\\
dying hair\\
fixing hair\\
getting a haircut\\
shaving head\\
shaving legs\\
trimming or shaving beard\\
washing hair\\
waxing back\\
waxing chest\\
waxing eyebrows\\
waxing legs\\

\noindent \textbf{hands (9)}\\
air drumming\\
applauding\\
clapping\\
cutting nails\\
doing nails\\
drumming fingers\\
finger snapping\\
pumping fist\\
washing hands\\

\noindent \textbf{head + mouth (17)}\\
balloon blowing\\
beatboxing\\
blowing nose\\
blowing out candles\\
brushing teeth\\
gargling\\
headbanging\\
headbutting\\
shaking head\\
singing\\
smoking\\
smoking hookah\\
sneezing\\
sniffing\\
sticking tongue out\\
whistling\\
yawning\\

\noindent \textbf{heights (15)}\\
abseiling\\
bungee jumping\\
climbing a rope\\
climbing ladder\\
climbing tree\\
diving cliff\\
ice climbing\\
jumping into pool\\
paragliding\\
rock climbing\\
skydiving\\
slacklining\\
springboard diving\\
swinging on something\\
trapezing\\

\noindent \textbf{interacting with animals (19)}\\
bee keeping\\
catching fish\\
feeding birds\\
feeding fish\\
feeding goats\\
grooming dog\\
grooming horse\\
holding snake\\
ice fishing\\
milking cow\\
petting animal (not cat)\\
petting cat\\
riding camel\\
riding elephant\\
riding mule\\
riding or walking with horse\\
shearing sheep\\
training dog\\
walking the dog\\

\noindent \textbf{juggling (6)}\\
contact juggling\\
hula hooping\\
juggling balls\\
juggling fire\\
juggling soccer ball\\
spinning poi\\

\noindent \textbf{makeup (5)}\\
applying cream\\
doing nails\\
dying hair\\
filling eyebrows\\
getting a tattoo\\

\noindent \textbf{martial arts (10)}\\
arm wrestling\\
capoeira\\
drop kicking\\
high kick\\
punching bag\\
punching person\\
side kick\\
sword fighting\\
tai chi\\
wrestling\\

\noindent \textbf{miscellaneous (9)}\\
digging\\
extinguishing fire\\
garbage collecting\\
laying bricks\\
moving furniture\\
spraying\\
stomping grapes\\
tapping pen\\
unloading truck\\

\noindent \textbf{mobility -- land (20)}\\
crawling baby\\
driving car\\
driving tractor\\
faceplanting\\
hoverboarding\\
jogging\\
motorcycling\\
parkour\\
pushing car\\
pushing cart\\
pushing wheelchair\\
riding a bike\\
riding mountain bike\\
riding scooter\\
riding unicycle\\
roller skating\\
running on treadmill\\
skateboarding\\
surfing crowd\\
using segway\\
waiting in line\\

\noindent \textbf{mobility -- water (10)}\\
crossing river\\
diving cliff\\
jumping into pool\\
scuba diving\\
snorkeling\\
springboard diving\\
swimming backstroke\\
swimming breast stroke\\
swimming butterfly stroke\\
water sliding\\

\noindent \textbf{music (29)}\\
beatboxing\\
busking\\
playing accordion\\
playing bagpipes\\
playing bass guitar\\
playing cello\\
playing clarinet\\
playing cymbals\\
playing didgeridoo\\
playing drums\\
playing flute\\
playing guitar\\
playing harmonica\\
playing harp\\
playing keyboard\\
playing organ\\
playing piano\\
playing recorder\\
playing saxophone\\
playing trombone\\
playing trumpet\\
playing ukulele\\
playing violin\\
playing xylophone\\
recording music\\
singing\\
strumming guitar\\
tapping guitar\\
whistling\\

\noindent \textbf{paper (12)}\\
bookbinding\\
counting money\\
folding napkins\\
folding paper\\
opening present\\
reading book\\
reading newspaper\\
ripping paper\\
shredding paper\\
unboxing\\
wrapping present\\
writing\\

\noindent \textbf{personal hygiene (6)}\\
brushing teeth\\
taking a shower\\
trimming or shaving beard\\
washing feet\\
washing hair\\
washing hands\\

\noindent \textbf{playing games (13)}\\
egg hunting\\
flying kite\\
hopscotch\\
playing cards\\
playing chess\\
playing monopoly\\
playing paintball\\
playing poker\\
riding mechanical bull\\
rock scissors paper\\
shuffling cards\\
skipping rope\\
tossing coin\\

\noindent \textbf{racquet + bat sports (8)}\\
catching or throwing baseball\\
catching or throwing softball\\
hitting baseball\\
hurling (sport)\\
playing badminton\\
playing cricket\\
playing squash or racquetball\\
playing tennis\\

\noindent \textbf{snow + ice (18)}\\
biking through snow\\
bobsledding\\
hockey stop\\
ice climbing\\
ice fishing\\
ice skating\\
making snowman\\
playing ice hockey\\
shoveling snow\\
ski jumping\\
skiing (not slalom or crosscountry)\\
skiing crosscountry\\
skiing slalom\\
sled dog racing\\
snowboarding\\
snowkiting\\
snowmobiling\\
tobogganing\\

\noindent \textbf{swimming (3)}\\
swimming backstroke\\
swimming breast stroke\\
swimming butterfly stroke\\

\noindent \textbf{touching person (11)}\\
carrying baby\\
hugging\\
kissing\\
massaging back\\
massaging feet\\
massaging legs\\
massaging person's head\\
shaking hands\\
slapping\\
tickling\\

\noindent \textbf{using tools (13)}\\
bending metal\\
blasting sand\\
building cabinet\\
building shed\\
changing oil\\
changing wheel\\
checking tires\\
plastering\\
pumping gas\\
sanding floor\\
sharpening knives\\
sharpening pencil\\
welding\\

\noindent \textbf{water sports (8)}\\
canoeing or kayaking\\
jetskiing\\
kitesurfing\\
parasailing\\
sailing\\
surfing water\\
water skiing\\
windsurfing\\

\noindent \textbf{waxing (4)}\\
waxing back\\
waxing chest\\
waxing eyebrows\\
waxing legs\\

\end{document}